\documentclass[letterpaper, 10 pt, conference]{ieeeconf}

\usepackage{cite}
\usepackage{graphicx}
\usepackage{tcolorbox}
\usepackage{xcolor}
\usepackage[hidelinks]{hyperref}
\usepackage[]{footmisc}
\usepackage{adjustbox}
\usepackage{booktabs,graphicx,makecell,siunitx,eqparbox}
\usepackage{tabularx}
\usepackage{footnote}
\usepackage{balance}
\usepackage{subfig}
\usepackage{caption}
\makesavenoteenv{tabular}
\usepackage{amssymb}

\usepackage{algorithm}
\usepackage{algorithmic}

\IEEEoverridecommandlockouts  
\usepackage[nopostdot,nogroupskip,nonumberlist]{glossaries}
\makeglossaries

\usepackage{array}
\newcolumntype{L}[1]{>{\raggedright\let\newline\\\arraybackslash\hspace{0pt}}m{#1}}
\newcolumntype{C}[1]{>{\centering\let\newline\\\arraybackslash\hspace{0pt}}m{#1}}
\newcolumntype{R}[1]{>{\raggedleft\let\newline\\\arraybackslash\hspace{0pt}}m{#1}}

\IEEEoverridecommandlockouts                              

\title{\LARGE \bf
DexGANGrasp: Dexterous Generative Adversarial Grasping Synthesis for Task-Oriented Manipulation
}

\author{Qian Feng$^{*,1,2}$, David S. Martinez Lema$^{*,1,2}$, Mohammadhossein Malmir$^{2}$, Hang Li$^{1,2}$,\\ Jianxiang Feng$^{1,2}$, Zhaopeng Chen$^{1}$, Alois Knoll$^{2}$
\thanks{*: Equal Contributions, \{qian.feng,  david.martinez\}@tum.de.}
\thanks{$^{1}$Agile Robots SE}
\thanks{$^{2}$ TUM School of Information Computation and Technology, Technical University of Munich}
}

\def\ie{\textit{i.e.}}

\def\GanModel{\textit{DexGanGrasp}}
\def\dexprompt{\textit{DexAfford-Prompt}}
\def\dexG{\textit{DexGenerator}}
\def\dexD{\textit{DexDiscriminator}}
\def\dexE{\textit{DexEvaluator}}

\def\graspVar{\mathbf{g}}
\def\pcdVar{\mathbf{x}}
\def\partPcdVar{\mathbf{x_{p}}}
\def\latentVar{\mathbf{z}}
\def\GeneratorParam{\theta}

\def\generatSet{\mathbf{G_{gen}}}
\def\generatGrasp{\mathbf{g_{gen}}}
\def\gtSet{\mathbf{G_{gt}}}
\def\partGrasp{\mathbf{g_{part}}}

\def\robotiq{Robotiq-3F hand}

\usepackage{booktabs}
\newcommand{\ra}[1]{\renewcommand{\arraystretch}{#1}}

\begin{document}

\maketitle

\newglossaryentry{bps}{name=BPS, description={Basis Point Set},first={Basis Point Set (BPS)}}
\newglossaryentry{cgm}{name=CGMs, description={Conditional Generative Models},first={Conditional Generative Models (CGMs)}}
\newglossaryentry{cvae}{name=cVAE, description={Conditional Variational Autoencoder},first={Conditional Variational Autoencoder (cVAE)}}
\newglossaryentry{ci}{name=CI, description={Mutual Cross-Information},first={Mutual Cross-Information (CI)}}
\newglossaryentry{cov}{name=Cov, description={Coverage},first={Coverage (Cov)}}
\newglossaryentry{dof}{name=DoF, description={Degrees of Freedom},first={Degrees of Freedom (DoF)}}
\newglossaryentry{dgm}{name=DGM, description={Deep Generative Model},first={Deep Generative Model (DGM)}}
\newglossaryentry{ee}{name=EE, description={End Effectors},first={End Effectors (EE)}}
\newglossaryentry{fpr}{name=FPR, description={False Positive Rate},first={False Positive Rate (FPR)}}
\newglossaryentry{gan}{name=GAN, description={Generative Adversarial Network},first={Generative Adversarial Networks (GAN)}}
\newglossaryentry{cgan}{name=cGAN, description={Conditional Generative Adversarial Network},first={Conditional Generative Adversarial Network (cGAN)}}
\newglossaryentry{nf}{name=NFs, description={Normalizing Flows},first={Normalizing Flows (NFs)}}
\newglossaryentry{cnf}{name=cNF, description={Conditional Normalizing Flow},first={Conditional Normalizing Flows (cNF)}}
\newglossaryentry{iid}{name=\textit{iid}, description={independent and identically distributed},first={independent and identically distributed (\textit{iid})}}
\newglossaryentry{kld}{name=KLD, description={Kullaback-Leibler Divergence},first={Kullaback-Leibler Divergence (KLD)}}
\newglossaryentry{ll}{name=LL, description={Log-Likelihood},first={Log-Likelihood (LL)}}
\newglossaryentry{lars}{name=LARS, description={Learned accept/reject sampling},first={Learned accept/reject sampling (LARS)}}
\newglossaryentry{llm}{name=LLM, description={Large Language Models},first={Large Language Models (LLM)}}

\newglossaryentry{mllm}{name=MLLM, description={Multimodal Large Language Models},first={Multimodal Large Language Models (MLLM)}}

\newglossaryentry{knn}{name=KNN, description={k-Nearest Neighbors},first={k-Nearest Neighbors (KNN)}}

\newglossaryentry{mle}{name=MLE, description={Maximum Likelihood Estimation},first={Maximum Likelihood Estimation (MLE)}}
\newglossaryentry{mlp}{name=MLP, description={Multi-Layer Perceptron},first={Multi-Layer Perceptron (MLP)}}
\newglossaryentry{vae}{name=VAE, description={Variational Autoencoder},first={Variational Autoencoder (VAE)}}
\newglossaryentry{vi}{name=VI, description={Variational Inference},first={Variational Inference (VI)}}
\newglossaryentry{vlm}{name=VLM, description={Vision Language Models},first={Vision Language Models (VLM)}}
\newglossaryentry{magd}{name=MAGD, description={Mean Absolute Grasp Deviation},first={Mean Absolute Grasp Deviation (MAGD)}}
\begin{abstract}

We introduce \GanModel, a dexterous grasp synthesis method that generates and evaluates grasps with a single view in real-time. \GanModel~comprises a \gls{cgan}-based DexGenerator to generate dexterous grasps and a discriminator-like DexEvalautor to assess the stability of these grasps. 
Extensive simulation and real-world experiments showcase the effectiveness of our proposed method, outperforming the baseline FFHNet with an $18.57\%$ higher success rate in real-world evaluation. 
To further achieve task-oriented grasping, we extend \GanModel~to \dexprompt, an open-vocabulary affordance grounding pipeline for dexterous grasping leveraging \gls{mllm} and \gls{vlm} with successful real-world deployments.
For the code and data, visit our \href{https://david-s-martinez.github.io/DexGANGrasp.io}{\textcolor{blue}{website}}.
\end{abstract}


\section{Introduction}
As a crucial prerequisite for manipulation, robotic grasping is a fundamental skill for robots to start interacting with our environment. 
Low-dimensional end effectors like 2-jaw grippers or suction cups have been extensively studied with significant progress~\cite{levine2018learning, mousavian20196, Liang_2019, zeng2022robotic}. 
However, 2-jaw grippers suffer from low dexterity, which limits their applicability for task-oriented manipulations~\cite{touchcode, hang2024dexfuncgrasp}.

Dexterous hands allow for a broader range of solutions to grasp an object. However, they also present the challenge of generating high-quality grasps within a high-dimensional search space.
\robotiq~is used in this work. A grasp for \robotiq~consists of 6 \gls{dof} grasp pose and 12 \gls{dof} joint configurations. 
To efficiently generate meaningful grasp configurations for dexterous hands, especially from a single view, many approaches adopt learning-based methods. These methods may require time-consuming shape completion~\cite{lundell2020multi,wei2022dvgg,dominik2022}, map visual input to a single unique grasp instead of many~\cite{schmidt2018grasping,Liu19}, predict grasp translation and rotation in a cascaded manner~\cite{xu2023unidexgrasp}, or require additional contact anchor as input~\cite{ugg2023}.

Dexterous hands are designed to mimic human functionality, and humans often grasp objects with a specific purpose in mind, and the method of grasping serves that purpose. For example, a human might grasp a hammer by the handle to use it but grasp the opposite side when handing it over. Different tasks may require completely different ways of grasping the same object. Most existing task-oriented grasping methods are either limited to 2-jaw grippers~\cite{tong2024ovalprompt, song2024learning6doffinegrainedgrasp} or rely on human-labeled task grasping datasets~\cite{task_gcn, touchcode, hang2024dexfuncgrasp}.


%
In this work, we propose \GanModel, a method to directly generate diverse dexterous grasps from a single view in real-time. \GanModel~consists of a \gls{cgan}-based \dexG~which generates various high-quality grasps given random samples from Gaussian distributions, and \dexE, which further predicts the grasp stability. Furthermore, we extend \GanModel~to task-oriented grasping by introducing \dexprompt, an open-vocabulary affordance grounding pipeline for dexterous grasping, leveraging \gls{mllm}s and \gls{vlm}s. The user-initiated affordance is fed into a \gls{mllm} together with the detected object name to identify which part of the object supports the task affordance. Subsequently, a \gls{vlm} segments the part based on the object part name. Finally, grasps generated by \GanModel~targeting the segmented part are filtered for execution.  

Our main contributions are:

\begin{itemize}
    \item We propose a grasp synthesis and evaluation pipeline using a \gls{cgan} for grasping unknown objects with a dexterous hand from a single view in real-time.
    \item We introduce a new synthetic grasping dataset for \robotiq~encompassing 115 objects, 2.1 million grasps along with 5750 different scenes. 
    \item We extend \GanModel~to a task-oriented grasping pipeline called \dexprompt, which leverages \gls{mllm}s and \gls{vlm}s for open-vocabulary affordance grounding with dexterous grasping.
    \item Extensive experiments in both simulation and real-world settings, along with a detailed ablation study, demonstrate that our method outperforms the \gls{cvae} baseline and proves effective when integrated into \dexprompt~for task-oriented grasping with user-initiated affordance.
\end{itemize}
\section{Related Work}
\subsection{Grasp Synthesis}
Dexterous grasp synthesis aims to generate robust grasps for robotic hands that can withstand external disturbances~\cite{newbury2023deep, 6672028}. 

\textbf{Analytical grasp synthesis} leverages hand-crafted geometric constraints, heuristics, and point cloud geometric features~\cite{Lei2017cshape, jeremy2016oneshot, Lu2019Reconstruct} to generate grasps. Analytical methods often generate less diverse grasp synthesis with prolonged run times, particularly when working with partial object point clouds.
To tackle partial views, many works~\cite{Varley2016Reconstruct, Avigal20203Dreconstruct} rely on shape completion to obtain the entire object model and then use GraspIt!~\cite{miller2004graspit} to generate grasps with grasp quality metrics. 
However, both shape completion and analytical grasp synthesis on high-\gls{dof} multi-fingered hands are still highly time-consuming.

\textbf{Learning-based grasp synthesis} immensely focuses on \gls{cvae}~\cite{mousavian20196, veres2017modeling, ffhnet, wei2022dvgg} and \gls{gan}~\cite{lundell2020multi, DDGC2021, patzelt2019conditional}, autoregressive~\cite{dominik2022} and diffusion models~\cite{barad2023graspldm}.

\textbf{\gls{cvae}} are firstly applied in grasp synthesis for two-jaw grippers~\cite{mousavian20196} and for multi-fingered hands~\cite{veres2017modeling, ffhnet, wei2022dvgg}. 
\cite{wei2022dvgg} proposes a pipeline of point cloud completion, \gls{cvae}-based grasp generation and grasp refinement, resulting in a long computation time. 
A more simple yet effective pipeline is~\cite{ffhnet}, which takes as input directly partial point cloud.


\textbf{\gls{gan}}s have been widely studied as generative models, especially in the field of image synthesis~\cite{mirza2014conditionalgan,karras2019stylegan,isola2018pixel2pixelgan}, sim2real transfer~\cite{rao2020rlcycleganrl, sim2real_assembly20222} and so on.
Several works adopt \gls{gan}s for multi-fingered grasp synthesis with RGB-D input data~\cite{lundell2020multi, DDGC2021}. 
Because of an integrated shape completion module, their grasping pipeline is slow, taking approximately 8 seconds. In contrast, our model predicts full grasps much faster. Similarly, \gls{gan}s are employed in \cite{patzelt2019conditional} to predict a 6D grasp pose along with one of four predefined grasp types from depth images. We apply a \gls{gan} directly on the point cloud to achieve fast run time while extending it further to task-oriented grasping. 


So far, only a tiny number of scholars have attended to auto-regressive models~\cite{dominik2022, humt2023combining}, latent diffusion models~\cite{barad2023graspldm} and \gls{nf}~\cite{xu2023unidexgrasp}.
The auto-regressive model in~\cite{dominik2022} requires shape completion beforehand~\cite{humt2023combining}, but our method works directly on single-view point cloud.
 A \gls{cnf}-based approach in~\cite{xu2023unidexgrasp} consists of two separate models predicting rotation and translation in a cascaded way, while no real-world evaluations are provided. 


\subsection{Task-Oriented Grasping}
Task-oriented grasping for unknown objects requires robots to grasp object-specific parts for specific tasks~\cite{lee2022virtual}. Several approaches require a task-labeled grasping dataset~\cite{task_gcn, touchcode, hang2024dexfuncgrasp} or leverage off-the-shelf vision-language models~\cite{lerftogo2023,tang2023graspgpt,Li2024ShapeGraspZT} because of their reasoning capability.

Graph Convolutional Network (GCN) is utilized in~\cite{task_gcn, atad2023efficient} to encode the relationship between objects and tasks. 
\cite{touchcode} introduces touch code to represent the contacts between the hand and object parts. They further train a model with an attention mechanism to predict the functional grasps. Their model requires manually annotated touch code and full knowledge of the object to predict the grasp. This idea is further extended in~\cite{wei2024funcgrasp} for refinement steps to achieve stable and human-like grasps.
To increase the grasping dataset quality, \cite{hang2024dexfuncgrasp} introduces a Real-Simulation Annotation System to generate a synthetic dataset with the help of real-world human demonstrations.
Our framework, inspired by \cite{zheng2024evaluating}, leveraging \gls{mllm}s and \gls{vlm}s, enables task-oriented grasping without needing a dedicated dataset.

\cite{lerftogo2023} takes as input scene point cloud and a task-oriented object part name and outputs a ranked grasps located in this part region. However, they require multiple views to reconstruct the scene, whereas our approach works on a single view. A novel grasp evaluator with \gls{llm} input is proposed to achieve zero-shot task-oriented grasping in~\cite{tang2023graspgpt}. Both approaches are applied to a 2-jaw gripper, which limits the manipulability of the following tasks. A convex decomposition approach is utilized in~\cite{Li2024ShapeGraspZT} to better determine the parts. However, without a dedicated grasp generation module, the grasps mainly end up with top-down grasps, limiting their manipulability afterward. Our proposed \dexprompt~is built on top of a diverse grasp generation module \GanModel~for more flexible 6D grasps.

\section{Problem Formulation}
This work aims to address the challenge of grasp synthesis for unknown objects with a multi-fingered hand, utilizing only a partial object point cloud $\pcdVar \in \mathbb{R}^{N\times3}$.
A grasp $\mathbf{g} \in \mathbb{R}^d$ is represented by the 12-DOF hand joint configuration $\boldsymbol{j} \in \mathbb{R}^{12}$ and the 6D palm pose $(\mathbf{R},\mathbf{t}) \in SE(3)$.
Grasping success for \robotiq~is defined as its ability to lift an object 20 cm above its resting position without any slippage.
Since one object can be grasped in infinite equally successful ways, the underlying relationship between $\pcdVar$ and $\graspVar$ is one-to-many instead of one-to-one or many-to-one. We apply generative models \gls{cgan} to tackle this problem.


Therefore, to formally define the problem, we assume an empirical dataset of $N$ objects with their $N_i$ corresponding possible grasps $\mathcal{D} = \{\pcdVar_i, \{\graspVar_{ik}\}_{k=1}^{N_i}\}_{i=1}^N$ drawn from the unknown underlying conditional distribution $p^*(\graspVar|\pcdVar)$.
We need to train a probabilistic generative model $p_{\GeneratorParam}(\graspVar|\pcdVar)$ parameterized by $\GeneratorParam$ to approximate $p^*(\graspVar|\pcdVar)$.
During inference, we can synthesize grasps and rank them to find and execute one optimal grasp.

\section{Methods}
\begin{figure*}[htbp]
\vspace{10pt}
    \centering
    \includegraphics[width=\textwidth]{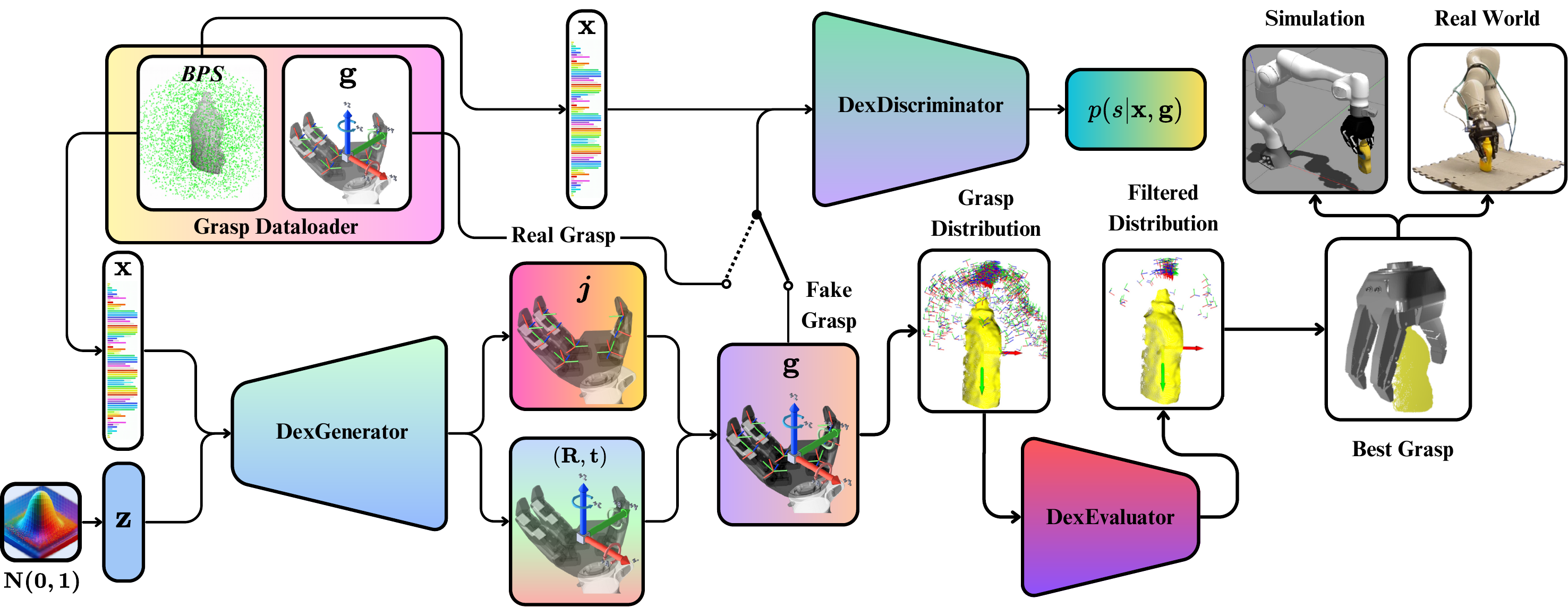}
    \vspace{5pt}
    \caption{\GanModel~consists of a \dexG, a \dexD~and a \dexE. \dexG~takes as input a bps-encoded point cloud as well as a sample $z$ from latent space. It tries to predict various "fake" grasps $\generatGrasp=(\mathbf{R},\mathbf{t},\mathbf{j})$, as close as possible to the positive ground truth "real" grasps $\graspVar_{pos}$. Meanwhile, \dexD~tries to differentiate between these two types of grasps. During inference, \dexE~filters stable grasps out of $\graspVar$ from \dexG~and the best grasp is chosen for the robot to execute.}
    \label{fig:ganmodel}
\end{figure*}

We propose a \GanModel~for dexterous grasp synthesis based on \gls{cgan}, shown in figure~\ref{fig:ganmodel}. The \GanModel~consists of a \dexG, a \dexD~and a \dexE.


\subsection{\dexG~and \dexD}
\label{sec:dexG}
The \dexG~$G$ takes the input of a noise sample from latent space $z \sim p_{z}(z)$ defined as a Standard Normal Gaussian $N (0, 1)$, and generates a synthetic grasp distribution $ \graspVar \sim p_{g}(z | \pcdVar)$ conditioned on point cloud $\pcdVar$. The \dexG~is trained to capture the ground truth grasp distribution $p_r$ and generate synthetic or "fake" generated grasps $\generatGrasp \sim G(z|\pcdVar)$ as realistically as possible.
On the other hand, the \dexD~$D$ tries to differentiate between the positive ground truth "real" grasp $ \graspVar_{pos} \sim p_{r}$ or a "fake" grasp $\generatGrasp$ by maximizing $\mathbb{E}_{\graspVar_{pos} \sim  p_{r}(\graspVar_{pos})} [\log{D(\graspVar_{pos})}]$ and minimizing $\mathbb{E}_{z \sim  p_{z}(z)} [\log{(1-D(G(z|\pcdVar)))}]$.

The training of \GanModel~consists of two steps. The first step updates the discriminator weights alone with discriminator loss $\mathcal{L}_{D}$ leaving the generator frozen, in eq.~\ref{eq:gen_loss}. 

\begin{align}
    \begin{split}
    \label{eq:gen_loss}
    \begin{aligned}
    \mathcal{L}_{D} & =  -\mathbb{E}_{\graspVar_{pos} \sim p_{r}(\graspVar_{pos})} [\log{(D(\graspVar_{pos})})] \\
    & - \mathbb{E}_{z \sim  p_{z}(z)} [\log{(1 - D(\generatGrasp))}]
    \end{aligned}
    \end{split}
    \end{align}

The second step will update the generator and discriminator weights together with generator loss $\mathcal{L}_{G}$.

\vspace{5pt}
The discriminator loss $\mathcal{L}_{D}$ and vanilla generator loss alone are not sufficient to train the \GanModel~model, due to the large diversity of the grasp distribution. Therefore, we define the generator's loss $\mathcal{L}_{G}$ as a combination of the adversarial loss $\mathcal{L}_{fake\_gen}$ and $\mathcal{L}_{dist}$, a weighted sum of $L2$ distance losses from translation, rotation, and joint configuration. $\mathcal{L}_{dist}$ guides the model into generating a grasp distribution close to the ground truth. The weights are determined manually to put all the losses on the same scale.

\begin{align}
   \mathcal{L}_{G} &=  L_{fake\_gen} + L_{dist}
    \\
    \mathcal{L}_{fake\_gen} &= \mathbb{E}_{z \sim  p_{z}(z)} [\log{(1-D(\generatGrasp ))}]
    \\
    \mathcal{L}_{dist} &= w_{t}L_{transl} +w_{r}L_{rot} + w_{j}L_{joint}
\end{align}

\subsection{\dexE}
It's intrinsically complicated to generate all equally successful grasps directly from a grasp distribution as a grasp is high-dimensional in a space with mixed manifolds. 
To secure the grasping success, we propose a \dexE~as an effective complementary option following the generator as it can model both positive $\graspVar_{pos}$ grasps and negative grasps $\graspVar_{neg}$ in a supervised manner. \dexE~has the same model architecture as \dexD~but it is trained separately, fed with both positive and negative grasps.

\dexE~predicts the probability $p(s | \graspVar, \pcdVar)$ of a grasp that results in a stable lift motion of the object, with a loss function given grasp label $y$:

\begin{equation}
    \label{eq:binary_crossentropy}
    \mathcal{L}_{Eva} = - \left[ y \log (f(\graspVar_{pos}, \pcdVar)) + (1 - y) \log (f(\graspVar_{neg}, \pcdVar)) \right]
\end{equation}

\subsection{\GanModel~at Inference}
When there is an unknown object point cloud denoted by $\pcdVar^*$, we can generate the corresponding grasps $\graspVar^*$ for this object by sampling from the latent space: 

\begin{equation}
    \graspVar^* \sim G(\graspVar|\latentVar, \pcdVar^*).
\end{equation}
Afterward, \dexE~predicts scores for all the generated grasps $\generatGrasp$, and the grasp with the best score will be chosen for the robot to execute. 

\begin{figure*}[ht]
    \vspace{10pt}
    \centering
    \includegraphics[width=\textwidth]{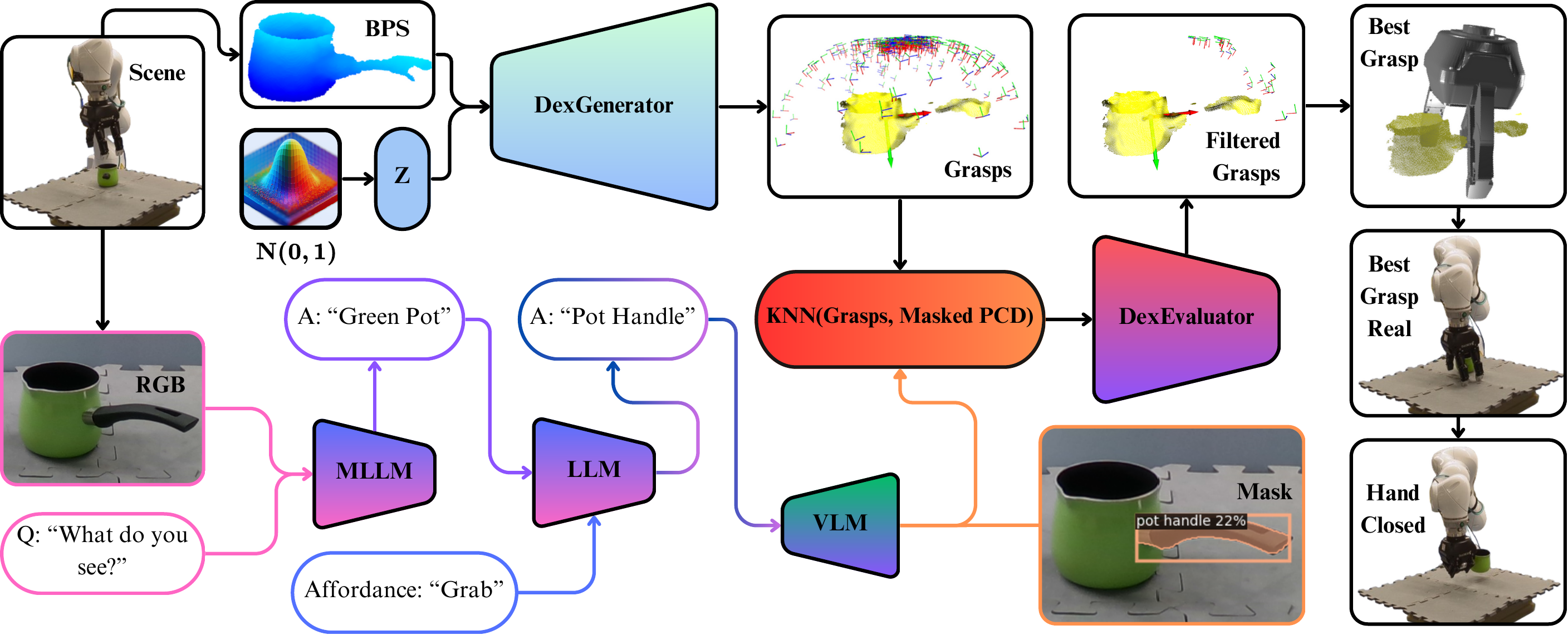}
    \caption{\dexprompt~builds on top of \GanModel~as an open-vocabulary affordance grounding pipeline to achieve task-oriented dexterous grasping. Firstly, an RGB image and the question "What do you see?" are fed to ChatGPT~4o (\gls{mllm})~\cite{openai2024chatgpt} to obtain the object name. The detailed prompt is in~\ref{block:affordance_prompt}. This object name, along with a user-initiated affordance such as "grab", is fed into ChatGPT~4 (\gls{llm})~\cite{openai2024chatgpt} to predict the object part name, which affords the task. Further,VLPart (\gls{vlm})~\cite{peize2023vlpart} segments the part from image space and projects to 3D space to obtain object part point cloud. A \gls{knn}-based filtering function is used to filter out the grasps that do not target the object part. Finally, the grasps are ranked with \dexE~for final execution. }
    \label{fig:dexprompt}
\end{figure*}
\subsection{\dexprompt}
We extend \GanModel~to a task-oriented grasping pipeline \dexprompt, described in Algorithm~\ref{alg:task-grasping}. This pipeline is inspired by~\cite{tong2024ovalprompt}, and we further simplify it by using \gls{mllm}s to replace the object detection part consisting of a \gls{vlm} and a list of object tokens in~\cite{tong2024ovalprompt}.
Our method takes as input an RGB image and uses the \gls{mllm} first to find what is the object on the table. Together with the object name and user-defined affordance, the \gls{llm} predicts the object part name, which corresponds to the affordance described in the following block.

\begin{tcolorbox}[title=Affordance Prompting with \gls{mllm}/\gls{llm}]
Q: Given an image, list the object name on the table with simple words such as cup, bottle, and book. Only answer the object's name. Don't say anything else."
\\
A: \textcolor{red}{[the object name]}
\\
Q: affordance is: \textcolor{red}{[given by user]}, please answer \textcolor{red}{[the object part name]} from the image based on the given affordance.
\\
A: \textcolor{red}{[the object part name]}
\label{block:affordance_prompt}
\end{tcolorbox}

The predicted object part name is then fed into the \gls{vlm} together with an RGB image to obtain the part segmentation.
Given part segmentation, which affords the task prompt, we project it to the point cloud to obtain the part point cloud. Afterwards, we apply \gls{knn} to filter generated grasp distribution. The filtering cost function is defined based on Euclidean and rotation distance between each grasp and object part point cloud, 
to ensure a grasp acting on the object part. Finally, this new list of grasps will be ranked with scores produced by \dexE~and the robot will execute a top-ranked grasp.

\begin{algorithm}
\caption{Task-Oriented Grasping Pipeline}
\label{alg:task-grasping}
\begin{algorithmic}[1]
\REQUIRE affordance, Object point cloud $\pcdVar$, object part point cloud $\partPcdVar$
\ENSURE top-ranked grasp

\STATE \textcolor{blue}{\# Affordance prompt}
\STATE $object\_name \leftarrow \gls{mllm}(image)$
\STATE $part\_name \leftarrow \gls{llm}(affordance, object\_name)$
\STATE $part\_segmentation \leftarrow \gls{vlm}(part\_name)$
\STATE $\partPcdVar \leftarrow crop(\pcdVar, part\_segmentation)$

\STATE \textcolor{blue}{\# Grasp synthesis and ranking}
\STATE $\generatGrasp \leftarrow \dexG(\latentVar,\pcdVar)$
\STATE $\partGrasp \leftarrow KNN(\partPcdVar, \generatGrasp)$
\STATE $filtered\_grasps \leftarrow \dexE(\partGrasp)$
\STATE $top\_grasp \leftarrow select\_best(filtered\_grasps)$
\RETURN $top\_grasp$
\end{algorithmic}
\end{algorithm}

\subsection{Implementation Details}
\label{sec:details}
The code is implemented with PyTorch 2.3.1.
All three proposed models are built on top of fully connected residual blocks with skip connections~\cite{he2016deep}. In $\mathcal{L}_{fake\_gen}$ we label all the $\generatGrasp$ generated grasps as real. This is done in order to prevent the \dexD~from easily identifying the generator's output since it is already trained with correctly labeled generated grasps during its update stage, defined by $\mathcal{L}_{D}$. This, in theory, should create a large gradient when \dexD~is presented with generated grasps, which would backpropagate through the whole network. This labeling approach is inspired by~\cite{gantrick}, where the labels are switched stochastically as a method to add noise to the discriminator to balance the adversarial training. \dexG~and \dexD~are both trained with learning rate of $1e-4$, batch size of 32 and 60 epochs. 
\dexE~is trained with lr of $1e-4$, batch size of 1024 and 30 epochs.

We use ChatGPT 4o~\cite{openai2024chatgpt} as \gls{mllm} for object detection, ChatGPT 4~\cite{openai2024chatgpt} as \gls{llm} for affordance prompting, and VLPart~\cite{peize2023vlpart} as \gls{vlm} for part detection. For \gls{knn}, we set k to 30 grasps.
\section{Experiments}
\subsection{Experimental Setup}
We conduct grasping experiments both in simulation and the real world to evaluate our proposed methods.

The experimental setup in the simulation consists of a Franka robot, \robotiq, and a RealSense D415 camera. In the real-world experiment, the robot arm is replaced by a Kuka iiwa, as our model is robot-independent. The robot grasping workspace is calibrated via homography~\cite{2023pickplace}.

We segment the point cloud from the RealSense camera using RANSAC plane removal to obtain the segmented object point cloud. This point cloud is encoded using \gls{bps}~\cite{prokudin2019efficient} and fed into \dexG~to generate grasp candidates. The top-ranked grasp by \dexE~is then executed by the robot.

\paragraph{Datasets}
\label{sec:heuristic_planner}
We generate a synthetic grasping dataset for \robotiq~in simulation Gazebo~\cite{koenig2004design}, following the setup in~\cite{ffhnet} for both training and evaluation. Specifically, a heuristic grasp planner is used to generate grasp data. The object models are sourced from the BIGBIRD~\cite{singh2014bigbird}, KIT~\cite{kasper2012kit}, and YCB~\cite{calli2015ycb} datasets.
For training in simulation, we use 115 graspable objects filtered from approximately 280 objects in the BIGBIRD and KIT datasets based on their graspability and object type. A total of 43,279 unique grasps have been executed in the simulation. With data augmentation of spawning objects in different poses, we eventually obtain 2.1 million grasps across 5750 various grasping scenes.

\begin{table*}[t]
\vspace{5pt}
\centering
\caption{Success Rate Comparison in Simulation}
\begin{center}
\label{tab:grasping_simulation_succ} 
\begin{adjustbox}{width=\textwidth}
\begin{tabular}{r|rrrrrrrrrrrrrrrr|r}
\toprule [1.2pt]
& \multicolumn{16}{c|}{Objects} & \multicolumn{1}{c}{} \\
Methods & \makecell{Bath \\ Detergent} & \makecell{Curry} & \makecell{Fizzy\\Tablets} & \makecell{Instant\\Sauce} & \makecell{Nut\\Candy} & \makecell{Potatoe\\Dumpling} & \makecell{Spray\\flask} & \makecell{Tomato\\Soup} & \makecell{Toy\\Airplane} & \makecell{Bleach\\Cleanser} & \makecell{Pitcher\\Base} & \makecell{Mug} & \makecell{Cracker\\Box} & \makecell{Mustard\\Bottle} & \makecell{Pudding\\Box} & \makecell{Mini\\Soccer \\Ball} & \makecell{Average \\ Succ \\Rate}  \\ [0.0001cm]
\midrule
FFHNet w/o eval & $95.0\%$ & $70.0\%$ & $85.0\%$ & $95.0\%$ &$95.0\%$& $95.0\%$ & $95.0\%$ & $90.0\%$ & $25.0\%$ & $80.0\%$ & $95.0\%$ & $95.0\%$ & $80.0\%$ & $65.0\%$ & $85.0\%$ & $100.0\%$ & $84.06\%$ \\
\GanModel~w/o eval  & $95.0\%$ & $80.0\%$ & $80.0\%$ & $90.0\%$ &$100.0\%$& $100.0\%$ & $100.0\%$ & $100.0\%$ & $50.0\%$ & $100.0\%$ & $95.0\%$ & $100.0\%$ & $100.0\%$ & $95.0\%$ & $80.0\%$ & $95.0\%$ & $91.25\%$ \\
\midrule
FFHNet with eval & $100.0\%$ & $90.0\%$ & $100.0\%$ & $100.0\%$ &$95.0\%$& $100.0\%$ & $100.0\%$ & $100.0\%$ & $15.0\%$ & $100.0\%$ & $90.0\%$ & $80.0\%$ & $70.0\%$ & $70.0\%$ & $100.0\%$ & $95.0\%$ & $87.81\%$ \\

\GanModel~with disc & $100.0\%$ &	$70.0\%$ &	$75.0\%$ &	$100.0\%$ &	$100.0\%$ &	$85.0\%$ &	$100.0\%$ &	$90.0\%$ & $40.0\%$ & $100.0\%$ &	$100.0\%$ &	$90.0\%$ &	$100.0\%$ & $95.0\%$ &	$100.0\%$ &	$100.0\%$ &	$90.31\%$ \\

\GanModel~with eval & $100.0\%$ & $100.0\%$ & $85.0\%$ & $100.0\%$ &$100.0\%$& $100.0\%$ & $100.0\%$ & $100.0\%$ & $45.0\%$ & $95.0\%$ & $100.0\%$ & $100.0\%$ & $100.0\%$ & $100.0\%$ & $95.0\%$ & $100.0\%$ & $\bold{95.0\%}$ \\
\end{tabular}
\end{adjustbox}
\end{center}
\end{table*}




\begin{table}[h]
\centering
\caption{Success Rate Comparison in Real World}
\begin{center}
\label{tab:grasping_real_world_succ} 
\begin{adjustbox}{width=\linewidth}
\begin{tabular}{r|rrrrrrr|r|r}
\toprule [1.2pt]
& \multicolumn{7}{c|}{Objects} & \multicolumn{1}{c|}{} \\
Methods & \makecell{Realsense\\Box} & \makecell{Mustard\\Bottle} & \makecell{Agile\\Bottle} & \makecell{Spam\\ Can} & \makecell{YCB \\Drill} & \makecell{Red\\Mug} & \makecell{Green\\Pan} & \makecell{Average \\ Succ \\ Rate} & \makecell{Inference \\ Speed} \\ [0.0001cm]
\midrule
FFHNet & $90.0\%$ & $70.0\%$ & $90.0\%$ & $80.0\%$ &$60.0\%$& $70.0\%$ & $50.0\%$ & $72.86\%$ &$30ms$ \\
\GanModel & $100.0\%$ & $100.0\%$ & $90.0\%$ & $90.0\%$ &$80.0\%$& $100.0\%$ & $80.0\%$ & $\bold{91.43\%}$ &$30ms$ \\
\end{tabular}
\end{adjustbox}
\end{center}
\end{table}

\begin{table}[h]
\centering
\vspace{10pt}
\caption{Task-Oriented Grasping Results in Real World}
\begin{center}
\label{tab:task_oriented_grasping_succ} 
\begin{adjustbox}{width=\linewidth}
\begin{tabular}{c|ccccc|c}
\toprule [1.2pt]
\makecell{Objects \\ (affordance)} & \makecell{Hammer\\(hand over)} & \makecell{Hammer\\(use)} & \makecell{Green Pan \\(grab)} & \makecell{Spray\\(use)} & \makecell{Red Brush \\(grab)} & \makecell{Average \\ Succ Rate}  \\ [0.0001cm]

\midrule
Grasping & $4/5$ & $2/5$ & $3/5$ & $4/5$ & $5/5$ & $72.0\%$ \\
\gls{vlm}~\cite{peize2023vlpart} & $6/6$ & $9/11$ & $6/9$ & $6/7$ & $5/9$ & $77.95\%$ \\
\vspace{-15pt}
\end{tabular}
\end{adjustbox}
\end{center}
\end{table}

\subsection{Evaluation Metrics}
To compare the quality of generated grasp distribution further, we propose two following metrics.
\textbf{\glsfirst{magd}}:
Built upon Mean Absolute Angular Deviation (MAAD)~\cite{Murphy2021implicitpdf}, our suggested \gls{magd} measures the `distance' between two sets of grasps, \ie, the generated set $\generatSet= \{\mathbf{g_{gen,i}}\}$ and the ground truth set $\gtSet = \{\mathbf{g_{gt,i}}\}$.
First, each grasp in $\generatSet$ finds the closest grasp in $\gtSet$ in terms of the L2 Euclidean distance of the translation $\mathbf{t}$.
\begin{equation}
\label{eq: grasp_distance}
   k = \arg \min_{i} \,  d( \mathbf{g_{gen,j}},\mathbf{g_{gt,i}}), \text{where }  d = \lVert \mathbf{t_{gen}} - \mathbf{t_{gt}} \rVert_2
\end{equation}
We exclude the rotation $\mathbf{R}$ and joint configuration $\mathbf{j}$ in this step because they are in a different order of magnitude to $\mathbf{t}$.
Therefore, we cannot simply sum them up but instead choose one of them. 
Second, we calculate the distances for the translation $\mathbf{t}$ and joint configuration $\mathbf{j}$ and geodesic distance for rotations $\mathbf{R}$.
For translations $\mathbf{d_{transl}}$ and joint configurations $\mathbf{d_{joint\_conf}}$ we calculate the distance in Euclidean space.
\begin{align} 
   \mathbf{d_{transl}} &=  \lVert \mathbf{t_{gen} - t_{gt,k}} \rVert \\ 
    \mathbf{d_{joint\_conf}} &= \lVert \mathbf{\theta_{gen} - \theta_{gt,k}} \rVert 
\end{align}
For rotations $\mathbf{R}$, we calculate the geodesic distance $\mathbf{d_{rot}}$:
\begin{align}
    \mathbf{R_{ref}} &= \mathbf{R_{gen}}^T \cdot \mathbf{R_{gt,k}} \nonumber \\ 
    cos(\mathbf{\alpha}) &= ( trace(\mathbf{R_{ref}}) -1)/2 \nonumber \\
    \mathbf{d_{rot}} &= \lvert arccos(cos(\mathbf{\alpha})) \lvert 
\end{align}
However, \gls{magd} only measures the quality of the best grasp in  $\generatSet$, hence less capable of measuring the diversity of the generated grasps. 
If the generated grasps are all identical and very close to one ground truth grasp, this will still have a very high \gls{magd} score.
For distances of translation, rotation, and joint configurations, we can calculate the per-grasp average L1 distance or the sum of the L2 distance for all grasps.

\textbf{\glsfirst{cov}}: 
It measures the fraction of grasps in the ground truth grasp set $\mathbf{G}_{gt}$ that is matched to at least one grasp in the generated set $\mathbf{G}_{gen}$: 
\begin{align}
    Cov = \frac{|\{
    \arg \min_{\mathbf{G}_{gt}} d(\mathbf{g}_{gen},\mathbf{g}_{gt}) |\mathbf{g}_{gen} \in \mathbf{G}_{gen} \}|} {|\mathbf{G}_{gt}|}
\end{align}
For each grasp in the generated set $\mathbf{G}_{gen}$, its nearest neighbor based on L2 distance in the
ground truth set $\mathbf{G}_{gt}$ is marked as a match.
\gls{cov} can be used to quantify the diversity of the generated grasp set with the ground truth set as reference.

\subsection{Simulation Grasping Results}

We conduct grasping experiments in simulation for 16 unknown objects, 8 KIT objects, and 8 YCB objects in table~\ref{tab:grasping_simulation_succ}. Each object is grasped 20 times, resulting in a total of 320 grasp trials. 

Firstly,  we evaluate the grasp quality alone from generative models as \gls{gan} in \GanModel~and \gls{vae} in FFHNet, without \dexE.
To facilitate a fair comparison for the grasp generator without a grasp evaluator, we evaluate the grasp samplers in simulation by executing the top 20 grasps instead of the single topmost one.
Here \GanModel~achieves a high success rate of $91.25\%$, surpassing FFHNet's $84.06\%$. With \dexE, \GanModel~still maintains a $7.19\%$ higher success rate than FFHNet. We observe fewer failures from \GanModel~in terms of unstable grasp poses due to wrong palm translation or rotation, indicating higher grasp quality both before and after filtering from \dexE.

We further compare the filtering capability of grasps with either \dexE~or \dexD. \dexE~outperforms \dexD~by $4.69\%$ while filtering \GanModel. This difference can be attributed to the fact that \dexD~is evaluating grasps based on realism (trained on $\graspVar_{pos}$ and $\generatGrasp$), while \dexE~is evaluating grasps based on probability of grasping success (trained on $\graspVar_{pos}$ and $\graspVar_{neg}$). This means the discriminator itself cannot be directly used as the evaluator as $\generatGrasp$ can be either positive or negative grasps.

\begin{figure}[t]
    \centering
    \includegraphics[width=\linewidth]{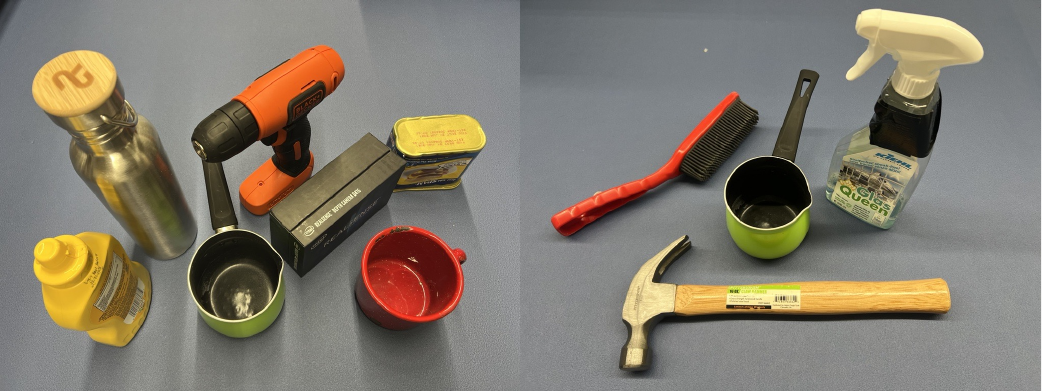}
    \vspace{5pt}
    \caption{On the left side, there are seven objects used for general grasping experiments: Agile Bottle, YCB Drill, RealSense Box, Spam Can (top row); Mustard Bottle, Green Pan, Red Mug (bottom row). On the right side, there are four objects specifically used for testing task-oriented grasping: Red Brush, Green Pan, Spray, and Hammer.}
    \label{fig:dexprompt}
\end{figure}

\subsection{Real world Grasping Results}
Following the simulation experiments, we conduct the real-world grasping experiment. We test 7 unknown objects with \GanModel~and FFHNet, both paired with \dexE. Each object is grasped 10 times, resulting in a total of 70 grasp trials for each model in table~\ref{tab:grasping_real_world_succ}.
In the real-world experiment, \GanModel~achieves the best success rate of $91.43\%$ outperforming FFHNet by $18.57\%$. For real-time performance, \GanModel~achieves an inference speed of $30ms$, tested in a commercial laptop with RTX 4090 16GB GDDR6 GPU.

\subsection{Failure Analysis}
Our model \GanModel~achieves high success rates across most objects, with particularly notable performance improvements over the baseline FFHNet. However, the model struggles with the Toy Airplane, Curry, and Fizzy Tablets (Sim) and  Green Pan, YCB Drill (Real), which have success rates of $40.0\%$, $70.0\%$, $75.0\%$, $80.0\%$, $80.0\%$ respectively. These items are challenging to grasp due to their irregular shapes or small graspable surfaces, which complicate secure grasp generation. While our model excels in reducing rotation, translation, and joint configuration errors, it exhibits a slight tendency to collide the palm with objects as it prioritizes generating a secure grasp, sometimes resulting in failure for more delicate or oddly shaped items.

Conversely, FFHNet produces more grasps that are positioned farther from the objects, leading to issues such as slippage or incorrect grasping. This is evident in the Toy Airplane, Mustard Bottle (Sim) and Green Pan, YCB Drill (Real), where FFHNet's success rates are $15.0\%$, $70.0\%$, $50.0\%$, and $70.0\%$ respectively, with \dexE. If not filtered, these distant grasps often fail to secure the object correctly, causing it to slip or not be grasped at all, highlighting a key area where \GanModel~provides a significant improvement.

\subsection{Real World Task-Oriented Grasping Results}
For the task-oriented grasping with \dexprompt, we choose 4 tool-related objects and test them with 3 different prompts `hand over', `use' and `grab'. Each object with a different prompt is tested 5 times, resulting in a total of 25 grasps, with details in table~\ref{tab:task_oriented_grasping_succ}

\dexprompt~achieves an overall grasping success rate of $72.0\%$. For the hammer, it enables two different prompts: `hand over', where the robot is supposed to grasp the top part with the predicted part name of `hammer head' and leave the handle for another person or robot to take over; and `use', where the robot is supposed to grasp the bottom handle with the predicted part name of `hammer handle'.

\begin{figure}[t]
    \centering
    \vspace{6pt}
    \includegraphics[width=\linewidth]{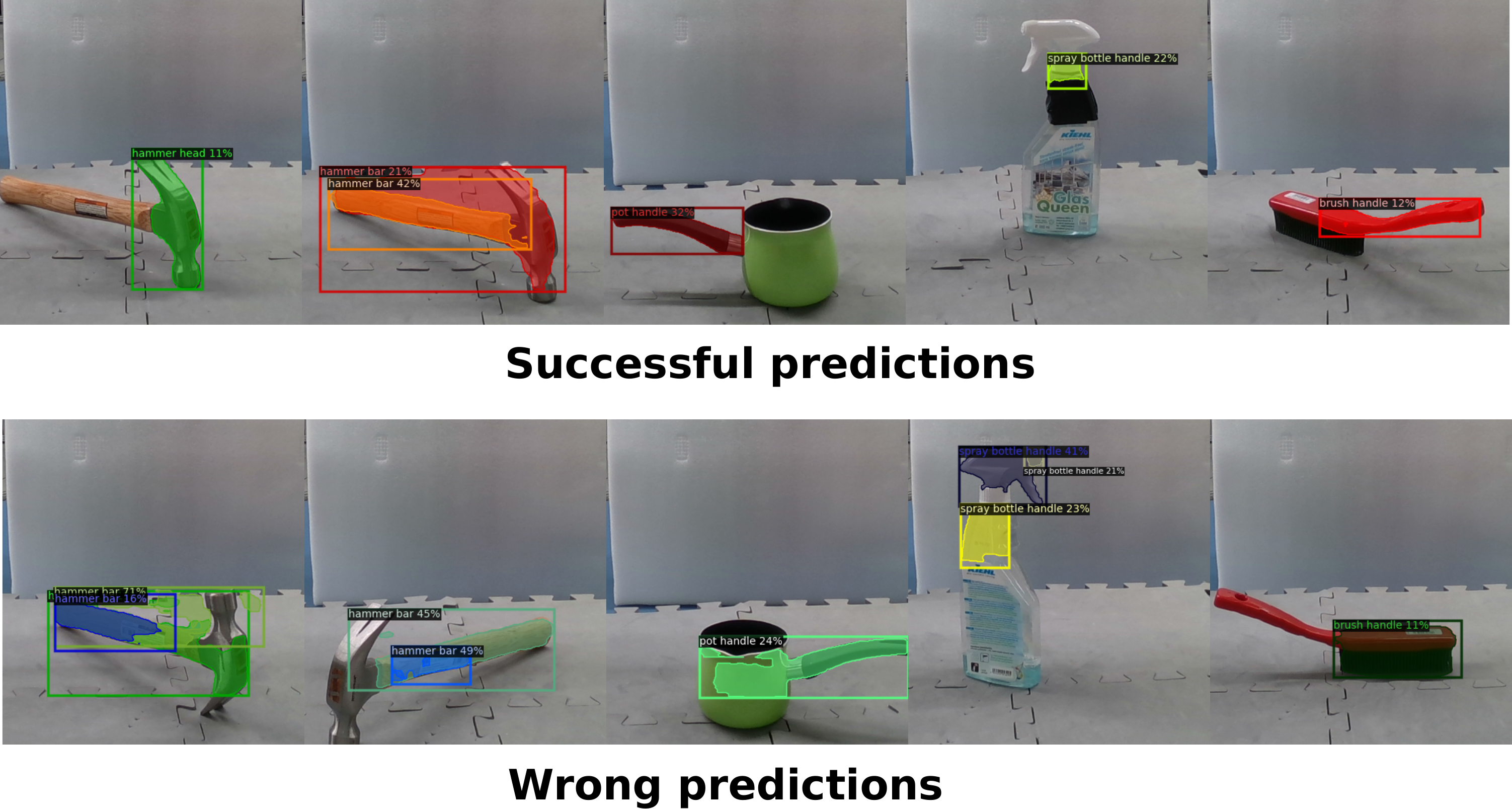}
    \caption{Successful part predictions on the first row and some failure cases in the second row from VLPart (\gls{vlm})~\cite{peize2023vlpart}.}
    \label{fig:vlpart}
\end{figure}

Besides grasping, the \gls{vlm} achieves an average success rate of $77.95\%$ for part segmentation in table~\ref{tab:task_oriented_grasping_succ}. The number of attempts for the \gls{vlm} are often more than 5 times because certain grasps are not executed successfully. For example, the robot goes into a singularity or reaches the joint limit. We observed several failures from the \gls{vlm}, which can occasionally predict the wrong part, detailed in figure~\ref{fig:vlpart}. The reasons could be the different lighting conditions compared to the training data in the \gls{vlm}.

\subsection{Ablation Study}

We perform an ablation study to assess various design choices and loss functions for our \GanModel compared to the baseline FFHNet (table~\ref{tab:ablation_study}). We test different architectures and hyperparameters, applying the \gls{magd} and \gls{cov} metrics over a random batch of 64 items and their grasp distributions.

The final \GanModel~ variant employs a \gls{cgan} with distance losses and reduced discriminator training (section~\ref{sec:details}). We identify two optimal checkpoints: Epoch 32, excelling in all metrics except \gls{cov}, and Epoch 37, which maximizes translation and coverage. We choose Epoch 32 because of its superior performance in key grasp metrics (translation, rotation, joint configuration).

We also test a vanilla \gls{cgan} without distance losses $\mathcal{L}{dist}$, which performs poorly. Adding $\mathcal{L}{dist}$ and increasing discriminator updates (omitting label-switching) improves results but lacks diversity in \gls{cov}. Next, we evaluate a Conditional Wasserstein \gls{gan} (cWGAN)\cite{NIPS2017_892c3b1c} using the Wasserstein distance as a loss. This approach is unstable, improving only \gls{cov}. Stabilizing with gradient penalties (cWGAN-GP\cite{NIPS2017_892c3b1c}) yields better results in coverage, rotation, and joint configuration, though translation errors persist.




\begin{table}[t]
\vspace{22pt}
\centering
\ra{1.2}
\caption{Metrics Analysis of \GanModel, its variations and baseline on \gls{magd} and \gls{cov}}\begin{center}
\label{tab:ablation_study} 
\begin{adjustbox}{width=\linewidth}
\begin{tabular}{c|ccc|c}
\toprule [1.35pt]
 & \multicolumn{3}{c|}{Cumulative MAGD(w/o eval)} & \multicolumn{1}{c}{}\\[0.15cm]
Ablated Models & Translation $ \downarrow$  & Rotation $\downarrow$ & Joints $\downarrow$ & Cov $\uparrow$\\
\midrule
FFHNet  & 30.22 & 23779.96 & 0.8621 & 29.47\% \\
			
\GanModel \ (ep = 37) & \textbf{25.75} & 25587.59 & 1.6773 & \textbf{29.77}\% \\
			
\GanModel \ (ep = 32) & 26.08& \textbf{22656.06} & \textbf{0.7249} & 25.91\% \\
		
Wasserstein \gls{cgan} with GP & 58.65 & 26496.25 & 0.9936 & 18.07\%\\
Wasserstein \gls{cgan}  & 71.10 & 38737.16 & 109.4223 & 1.04\% \\
\gls{cgan} (increased train rate of discriminator) & 46.52 & 37683.35 & 5.5801 & 0.88\%\\
Vanilla \gls{cgan} (without distance losses) & 166.49 & 38879.15 & 320.90 & 0.47\%\\
			
\end{tabular}
\end{adjustbox}
\vspace{-20pt}
\end{center}
\end{table}

\section{Conclusion}

In this work, we propose \GanModel~for the dexterous grasping problem utilizing \gls{cgan} to grasp unknown objects with a single view in real-time. The proposed \dexG~in \GanModel~demonstrates its ability to generate higher quality and more diverse grasps compared to baseline FFHNet, assessed through our proposed \gls{magd} and \gls{cov} metrics and extensive simulation as well as real-world evaluations. To address task-oriented grasping, we further extend \GanModel~to \dexprompt, enabling an open-vocabulary affordance grounding pipeline with successful real-world deployments to unknown objects, showcasing the potential for future advancements in task-based manipulation.

We have identified several limitations in the current pipeline. Firstly, when dealing with slightly transparent objects such as sprays, the point cloud from the RealSense camera can be very noisy on the surface, leading to an unrealistic grasp distribution from \GanModel. Secondly, the current state-of-the-art \gls{vlm} needs to be more stable for affordance prediction. An affordance-aware grasp generation model is a promising future direction.







\bibliographystyle{IEEEtran}
{\scriptsize
\vspace{0.01 cm}
\bibliography{bibliography}
}

\end{document}